\pdfoutput=1

\documentclass[11pt]{article}

\usepackage[]{latex/acl}

\usepackage{times}
\usepackage{latexsym}

\usepackage[T1]{fontenc}

\usepackage[utf8]{inputenc}

\usepackage{microtype}

\usepackage{inconsolata}

\usepackage{graphicx}

\usepackage{graphicx}
\usepackage{inconsolata}
\usepackage{microtype}
\usepackage{booktabs}
\usepackage{graphicx}
\usepackage{bm}
\usepackage{times}
\usepackage{latexsym}
\usepackage{amsmath,amssymb,amsfonts}
\usepackage{mathtools}
\usepackage{subcaption}
\usepackage{multirow}
\usepackage{booktabs}
\usepackage{xcolor}
\usepackage[switch]{lineno} 
\usepackage{stmaryrd}
\usepackage{amsthm}
\usepackage{paralist}

\usepackage{bbding}
\usepackage{algpseudocode}
\usepackage[ruled,noend,linesnumbered]{algorithm2e}
\usepackage{dcolumn}

\usepackage{soul}
\usepackage{xcolor}

\usepackage{fancyhdr}

\fancypagestyle{firstpage}{%
  \fancyhf{} 
  \fancyfoot[C]{In \textit{Findings of the Association for Computational Linguistics: EMNLP 2024}, pages 11572--11583}
}

\usepackage{stmaryrd}
\SetSymbolFont{stmry}{bold}{U}{stmry}{m}{n}
\definecolor{amaranth}{rgb}{0.8, 0.0, 0.}
\title{A Decoding Algorithm for Length-Control Summarization  \\
Based on Directed Acyclic Transformers
}

\newcolumntype{d}[1]{D{.}{.}{#1}}
\newcolumntype{g}[1]{D{.}{\%}{#1}}  
\newcommand{\topK}[1]{$\operatorname{top}$-$#1$}

\author{
  \textbf{Chenyang Huang}\textsuperscript{*\rm 1}, 
  ~\textbf{Hao Zhou}\textsuperscript{\textdagger \rm 2},   
  ~\textbf{Cameron Jen}\textsuperscript{\rm 1}, 
  ~\textbf{Kangjie Zheng}\textsuperscript{\rm 2}, \\
  ~\textbf{Osmar R. Za\"iane}\textsuperscript{\ddag \rm 1}, 
  ~\textbf{Lili Mou}\textsuperscript{\ddag \rm 1} \\
    \textsuperscript{\rm 1}\normalsize{Dept. of Computing Science, Alberta Machine Intelligence Institute (Amii), University of Alberta}\\
    \textsuperscript{\rm 2}\normalsize{Institute for AI Industry Research (AIR), Tsinghua University} ~ \\
  \tt{\normalsize{chenyangh@ualberta.ca}} ~ \tt{\normalsize{zhouhao@air.tsinghua.edu.cn}} ~ \tt{\normalsize{cjen@ualberta.ca}}  \\
   \tt{\normalsize{kangjie.zheng@gmail.com}} ~ \tt{\normalsize{zaiane@ualberta.ca}} ~ {\tt \normalsize{doublepower.mou@gmail.com}} 
}

\begin{document}
\maketitle
\begingroup\def\thefootnote{*}\footnotetext{Work partially done during an internship at AIR}\endgroup
\begingroup\def\thefootnote{\textdagger}\footnotetext{Corresponding author}\endgroup
\begingroup\def\thefootnote{\ddag}\footnotetext{Canada CIFAR AI Chair}\endgroup
\begin{abstract}
Length-control summarization aims to condense long texts into a short one within a certain length limit. Previous approaches often use autoregressive (AR) models and treat the length requirement as a soft constraint, which may not always be satisfied. 
In this study, we propose a novel length-control decoding algorithm based on the Directed Acyclic Transformer (DAT). Our approach allows for multiple plausible sequence fragments and predicts a \emph{path} to connect them. In addition, we propose a Sequence Maximum a Posteriori (SeqMAP) decoding algorithm that marginalizes different possible paths and finds the most probable summary satisfying the length budget. Our algorithm is based on beam search, which further facilitates a reranker for performance improvement. Experimental results on the Gigaword and DUC2004 datasets demonstrate our state-of-the-art performance for length-control summarization.\footnote{Our code, output, and data are available at \url{https://github.com/MANGA-UOFA/DAT-LenC} \label{foot:link}}
\end{abstract}

\thispagestyle{firstpage}

\section{Introduction}

Summarization systems aim to condense lengthy text into a shorter form, while preserving key information \cite{nenkova-etal-2011-automatic,rush-etal-2015-neural,schumann-etal-2020-discrete,tsvigun-etal-2022-active}. Recent studies underscore the importance of length-control summarization \cite{liu-etal-2018-controlling,takase-okazaki-2019-positional,liu-etal-2022-learning}. For example, the Google search engine limits webpage title displays to 63 characters, tweets have a 280-character cap, and scientific paper abstracts are typically restricted to a few hundred words.

Most previous text summarization methods rely on autoregressive (AR) models \cite{SutskeverVL14,attentionisallyouneed}, which generate a summary word by word. For length control, AR models typically treat the summary length as an additional signal during both training and inference \cite{liu-etal-2018-controlling,takase-okazaki-2019-positional}. However, such methods do not strictly control the generation length, and thus fall into the \emph{soft-constraint} category.

Recently, non-autoregressive (NAR) models \cite{gu2018nonautoregressive,lee-etal-2018-deterministic} have been applied to text summarization \cite{su-etal-2021-non,liu-etal-2022-learning}. Unlike its autoregressive counterpart, an NAR model predicts target words independently so that the length-control text generation problem can be decomposed into shared sub-problems, facilitating dynamic programming for efficient computation~\cite{liu-etal-2022-learning, liucharacter}. Such methods fall into the \emph{hard-constraint} category.

Specifically, \citet{liu-etal-2022-learning,liucharacter} apply the Connectionist Temporal Classification (CTC) model \cite[][]{graves2006connectionist}, which allows placeholders and repetitions during generation but removes them in a post hoc manner. However, these CTC-based methods tend to produce word-level extractive summaries due to their restricted modeling capacity. As a result, the generated summaries contain very few new words and mostly preserve the original word order \cite{chuang-etal-2021-investigating,shao2022}.

In this work, our first insight is to apply the Directed Acyclic Transformer \cite[DAT,][]{huang2022directed} for length-control summarization. DAT, originally introduced for machine translation, is a powerful NAR approach that expands its canvas to allow multiple plausible {text} fragments and predicts \emph{links} to connect them (the links along with the predicted words are called a \emph{path}). In this way, DAT is more flexible in predicting words and determining their order, resulting in better generation quality than CTC \cite{huang2022directed,huang2023directed}.

For length-control decoding based on DAT, we propose a Sequence Maximum a Posteriori \mbox{(SeqMAP)} decoding objective, which marginalizes possible linked steps and seeks the most probable sequence satisfying the length budget. Our SeqMAP is different from the existing MAP decoding objective \cite{shao-etal-2022-viterbi} that seeks the most probable path (links and words), which we refer to as PathMAP.
Since DAT training considers the marginalization of all possible links for a groundtruth sequence, our SeqMAP aligns with the training objective better (as it huggalso performs marginalization), and is expected to surpass PathMAP in length-control summarization. 

Compared with PathMAP, our SeqMAP is a more challenging decoding objective. The traditional PathMAP performs $\arg\max$ for both links and words, which can be reorganized according to prediction time steps and accomplished by dynamic programming in one pass. On the contrary, our SeqMAP performs $\arg\max$ for word selection after summing over all possible linked steps, but $\max$ and $\operatorname{sum}$ operations cannot be directly swapped to decompose the {overall}  problem by time steps, breaking down the dynamic programming algorithm.

To this end, we propose an approximate algorithm to address SeqMAP, where we nevertheless decompose the objective by time steps. For each step, we make a greedy selection for the $\max$ operation, where we only consider a beam of high-probability sequences for the sum operation. Further, we may apply a reranker by a pretrained BERT model \cite{zhuang-etal-2021-robustly} to select the best sequence in the beam for performance improvement. 

We perform extensive experiments using the Gigaword \cite{graff2003english} and DUC2004 \cite{bommasani-cardie-2020-intrinsic} datasets, following previous work in non-autoregressive length-control summarization \cite{liu-etal-2022-learning}.
Results show that both SeqMAP and PathMAP outperform existing models based on CTC, which justifies our choice of the DAT model. Further, our approximate algorithm for SeqMAP outperforms PathMAP even without the reranker, suggesting the superiority of the SeqMAP objective. Our reranker further improves the summarization quality consistently, demonstrating its effectiveness. 

\section{Methodology}
In this section, we first present the Directed Acyclic Transformer (DAT) and the existing PathMAP decoding method (Subsection~\ref{sec:dat}). 
Then, we introduce our SeqMAP decoding objective, and develop a beam search-based dynamic programming algorithm to approach it (Subsection~\ref{sec:beam}). Finally, we design a reranking process to select the best beam candidate (Subsection~\ref{sec:reranker}).

\subsection{Length Control with DAT}
\label{sec:dat}
\textbf{DAT Model.} Our first contribution is the adaptation of the Directed Acyclic Transformer \cite[DAT,][]{huang2022directed} to the length-control summarization task. DAT is a non-autoregressive (NAR) model, capable of generating multiple plausible output segments, which are then selected and connected via \emph{links} to form the final output. DAT provides more flexibility in word selection and ordering, compared with the existing NAR summarization system \cite{liu-etal-2022-learning}, which is based on the Connectionist Temporal Classification (CTC) model \cite{graves2006connectionist}.

Consider the source text $\mathbf x = (\mathrm x_1, \cdots, \mathrm x_{T_{\mathrm x}})$ and the corresponding groundtruth summary $\mathbf y = (\mathrm y_1, \cdots, \mathrm y_{T_{\mathrm y}})$, where $T_{\mathrm{x}}$ and $T_{\mathrm{y}}$ denote their lengths. 
To allow multiple plausible output segments, DAT expands its generation canvas by having $S$ prediction steps, typically $S \ge T_{\mathrm y}$.

At each step $s \in \{1, \cdots, S\}$, DAT makes a word prediction $p_{\text{word}}^{(s)}(\cdot)$ and a link  prediction $p_{\text{link}}^{(s)}(\cdot)$.

In particular, the word prediction is given by
\begin{align}
    p_{\text{word}}^{(s)}(\cdot| \textbf x) = \operatorname{softmax}(\mathbf W_{\text{word}} \bm h_{s}) \label{eq:word_prediction}
\end{align}
where $\mathbf W_{\text{word}} \in \mathbb R^{|\mathcal V| \times d}$ is a learnable matrix and $\bm h_{s} \in \mathbb R^{d}$ is the decoder's hidden state at step $s$, with $d$ being the dimension and $|\mathcal  V|$ representing the vocabulary size.

Link prediction forecasts the step of the subsequent word that follows the $s$th step:
\begin{align}
\begin{aligned}
    p_{\text{link}}^{(s)} (\cdot|\mathbf x) = \operatorname{softmax}([\bm k_{s}^{\top} \bm q_{{s}+1}; \cdots; \bm k_{s}^{\top} \bm q_{{S}}])  \label{eq:pos_prediction}
\end{aligned} 
\end{align}
where $\bm k_{s} = \mathbf W_{\mathrm k} \bm h_{s}$ and $\bm q_{s} = \mathbf W_{\mathrm q} \bm h_{s}$  are transformations of the hidden state, both $\mathbf{W}_{\mathrm{k}}$ and $\mathbf{W}_{\mathrm{q}}$ being learnable matrices in $\mathbb{R}^{d \times d}$. The $[;]$ operation concatenates scalars into a column vector.

Given a reference summary $\mathbf{y}$ of length $T_{\mathrm{y}}$ from the training set, the DAT model predicts links to connect $T_{\mathrm{y}}$-many steps among a total of $S$ generation steps.
We denote the linked steps by $\bm{a} = (a_1, \cdots, a_{T_{\mathrm{y}}})$, where $1 = a_1 < \cdots < a_{T_{\mathrm{y}}} = S$. Following \newcite{huang2022directed}, we refer to the linked steps and corresponding words as a \emph{path}.\footnote{In practice, two special tokens, $\langle\text{bos}\rangle$ and $\langle\text{eos}\rangle$, are added at the beginning and end of $\mathbf{y}$, respectively. {Therefore, every} path has $\langle\text{bos}\rangle$ at step $1$ and $\langle\text{eos}\rangle$ at step $S$.} 

The probability of a path---having the linked steps $\bm a$ and yielding the target summary $\mathbf{y}$---is given by
\begin{align}
\resizebox{.85\linewidth}{!}{$
\begin{aligned}
    p(\mathbf y, \bm a | \textbf x) = \prod_{t=2}^{T_{\mathrm{y}}} p_{\text{link}}^{(a_{t-1})}(a_t) \prod_{t=1}^{T_{\mathrm y}} p_{\text{word}}^{(a_t)}(\mathrm y_t) \label{eq:path_prob}
\end{aligned}
$}
\end{align}
where $p_{\text{link}}^{(a_{t-1})}(a_t)$ is the probability of linking the $a_{t-1}$th generation step to the $a_{t}$th, and $p_{\text{word}}^{(a_t)}(\mathrm{y}_t)$ is the probability of predicting the word $\mathrm{y}_t$ at the $a_t$th step. Conditioning on $\textbf x$ is omitted in $p_{\text{word}}$ and $p_{\text{link}}$ for brevity. Further denoting them by $l_{a_{t-1}, a_t}$ and $w_{a_t, \mathrm y_t}$, we rewrite Eqn.~\eqref{eq:path_prob} as
\begin{align}
    p(\mathbf y, \bm a | \textbf x) & = \prod_{t=2}^{T_{\mathrm{y}}} l_{a_{t-1}, a_t} \prod_{t=1}^{T_{\mathrm y}} w_{a_t, \mathrm y_t} \\
    & = w_{1,\mathrm y_1} \prod_{t=2}^{T_{\mathrm{y}}} l_{a_{t-1}, a_t}  w_{a_t, \mathrm y_t}
\end{align}

For DAT, obtaining the probability of generating a target summary $\mathbf{y}$ requires marginalizing over all possible sequences of linked steps, given by
\begin{align}
    p(\mathbf y | \textbf x) &= \sum_{\boldsymbol a \in \Gamma_{T_{\mathrm y},S} } p(\mathbf y, \boldsymbol a | \textbf x) \\
    & = \sum_{\boldsymbol a \in \Gamma_{T_{\mathrm y},S} } w_{1,\mathrm y_1} 
 \prod_{t=2}^{T_{\mathrm{y}}} l_{a_{t-1}, a_t} w_{a_t, \mathrm y_t} \label{eq:dag2}
\end{align}
where $\Gamma_{T_{\mathrm{y}},S} = \{ \bm a = ( a_1, \cdots, a_{T_{\mathrm y}}) | 1 = a_1 < \cdots < a_{T_{\mathrm{y}}} = S \}$ is the set of all possible $T_{\mathrm y}$-many linked steps among $S$-many generation steps.

Although direct enumeration over $\Gamma_{T_{\mathrm{y}},S}$ is intractable, a dynamic programming algorithm can efficiently compute the marginalization for training \cite{huang2022directed}.

\textbf{PathMAP Decoding.}
Recently, \newcite{shao-etal-2022-viterbi} propose a DAT-based decoding algorithm (referred to as PathMAP) that finds the most probable path of words and linked steps, given a specific length.

Formally, we consider a length-$T$ path of linked steps $\bm{a} \in \Gamma_{T,s}$ and predicted words $\mathbf{v}_{\bm{a}} = (\mathrm v_{a_1}, \cdots, \mathrm v_{a_{T}})$, where $\mathrm v_{a_t} \in \mathcal V$ is the predicted word at the $a_t$th step and $s \in \{T, \cdots, S\}$ is any valid prediction step allowing $T$ words. The most probable length-$T$ path is obtained by maximizing the joint probability of the linked steps and word predictions at any valid prediction step, given by
\begin{align}
    & \max_{s \in \{T, \cdots, S\} } \underset{\substack{\bm{a} \in \Gamma_{T,s} \\ \mathbf{v}_{\bm{a}} \in \mathcal V^{T}}}{\max} w_{1,\mathrm v_1} \prod_{t=2}^{T} l_{a_{t-1}, a_t}  w_{a_t, \mathrm v_{a_t}} \label{eq:path_map_1} \\ 
    = & \max_{s \in \{T, \cdots, S\} } \underset{\substack{\bm{a} \in \Gamma_{T,s} }}{\max} w_{1, \mathrm v_1^*} \prod_{t=2}^{T} l_{a_{t-1}, a_t}  w_{a_t, \mathrm v_{a_t}^*} \label{eq:path_map_2} \\ 
    = &  ~ \max_{s \in \{T, \cdots, S\} }  w_{s,\mathrm v_{s}^*} \max_{s^\prime \in \{T-1, \cdots, s-1\}} \Bigg\{ \nonumber \\ 
    & l_{s^\prime, S}  \underset{\bm{a} \in \Gamma_{T-1,s'}} {\max}  \Big( w_{1, \mathrm v_{1}^*} 
  \prod_{t=2}^{T-1} l_{a_{t-1}, a_t} ~  w_{a_{t}, \mathrm v_{a_{t}}^*} \Big)  \Bigg\} \label{eq:path_map_3}
\end{align}
In Eqn.~\eqref{eq:path_map_2}, we choose word predictions greedily because the $\max$ operation of a word is independent of the $\max$ operations over other words and linked steps. 
Further, Eqn.~\eqref{eq:path_map_2} can be decomposed into Eqn.~\eqref{eq:path_map_3} in a recursive fashion, allowing for efficient dynamic programming.

\subsection{Our SeqMAP Decoding}  
\label{sec:beam}
\textbf{A Limitation of PathMAP.}
As seen, the PathMAP objective described in Eqn.~\eqref{eq:path_map_1} performs $\max$ for both links and words, which is different from DAT's marginalization training objective. 
This is not ideal, as a discrepancy between training and inference often leads to performance degradation \cite{bengio2015scheduled,zhang-etal-2019-bridging}. 

\textbf{SeqMAP Objective.}
To this end, we propose a novel Sequence Maximum a Posteriori (SeqMAP) objective that marginalizes all possible linked steps to find the most probable sequence of length $T$. This is given by
\begin{equation}
    \mathbf y^*  \!\! = \underset{\substack{\mathbf y \in \mathcal V^{T}  } } {\arg\max} \!\!\!\!\! \sum_{ s \in \{T, \cdots, S\}} \sum_{\bm a \in \Gamma_{T,s}} \!\!\! w_{1,\mathrm y_1} \! \prod_{t=2}^{T} l_{a_{t-1},a_{t}} w_{a_t,\mathrm y_t} \label{eq:seqmap_obj}
\end{equation}

However, solving Eqn.~\eqref{eq:seqmap_obj} is challenging. This is because the $\arg\max$ and summation cannot be swapped to decompose the overall objective based on the sentence length, making it infeasible to design a SeqMAP-like dynamic programming algorithm.

\textbf{Decoding Algorithm for SeqMAP.}
We propose an approximate algorithm to maximize our SeqMAP objective. The general idea is to perform dynamic programming (DP) by re-organizing $\arg\max$ and summation operations based on the output length anyway, and further improve the efficiency of the summation with beam search.

Let $\mathcal{A}_{t,s}$ be the (approximated) top-$K$ length-$t$ sequences that are generated \emph{at or before} DAT's $s$th step, denoted by
\begin{align}
    \mathcal{A}_{t,s} & = \{\mathbf  b^{(k)}\}_{k=1}^{K} \label{eq:beam_a}
\end{align}
In addition, we need to store the probability of generating $\mathbf b^{(k)}$ ending at step $s'$; we denote it by $u_{s'}(\mathbf b^{(k)})$, for $t\le s'\le s$. This is because a sequence $\mathbf b^{(k)}$ can be generated at different steps before the $s$th, and tracking their probabilities is helpful for marginalization.

The initialization of $\mathcal{A}_{t,s}$ fills in the DP table for $t=0$, where each $\mathcal{A}_{0,s}$ (for $s > 0$) contains a special $\langle\text{bos}\rangle$ token, indicating the beginning of a sentence. Additionally, the score for each $u_s(\langle\text{bos}\rangle)$ is initialized to $1$.

The DP recursion iteratively computes $\mathcal{A}_{t,s}$ for every $t >0$ and every $s$ such that $t \le s \le S$. We observe that each $\mathcal{A}_{t,s}$ can be obtained by 1) expanding the length-$(t-1)$ sequences in $\mathcal{A}_{t-1,s-1}$ with words from the $s$th step, which results in sequences ending at step $s$, and 2) further merging them with the length-$t$ sequences in $\mathcal{A}_{t,s-1}$, which approximates the best length-$t$ sequences ending before step $s$. Therefore, we design a recursive step with an \textsc{Expand} operation and a \textsc{Merge} operation.

\textbf{\textsc{Expand.}} Let $\mathcal{B}_{t,s}$ be the approximated top-$K$ length-$t$ sequences that are generated \emph{exactly at} the $s$th step.
Intuitively, we can approximate $\mathcal{B}_{t,s}$ by expanding sequences in $\mathcal{A}_{t-1,s-1}$ with the word predictions at step $s$, given by
\begin{align}
    \mathcal{B}_{t,s} = \operatorname{top-}\!K \Big\{ \mathbf b \oplus \mathrm v  \,\Big| \, \mathbf b \in \mathcal{A}_{t-1,s-1},  \mathrm v \in \mathcal V \Big\} \label{eq:expand_only}
\end{align}
When expanding a partial sequence with $\mathrm v \in \mathcal V$, our implementation only considers the top-$V$ words based on the predicted {word} probability in Eqn.~\eqref{eq:word_prediction} to improve efficiency.

For each expanded sequence in $\mathcal{B}_{t,s}$, the corresponding scores are given by 
\begin{align}
    u_s(\mathbf b \oplus  \mathrm v) = w_{s,\mathrm v} \, \sum_{s'=t-1}^{s-1} \,  u_{s'}(\mathbf b) \cdot l_{s',s} \label{eq:expand_score}
\end{align}
Here, we marginalize out different previous linked steps that may generate $\mathbf b$, which follows the spirit of the marginalization in our SeqMAP objective~\eqref{eq:seqmap_obj}.
 
\textbf{\textsc{Merge.}}
By definition, $\mathcal{A}_{t,s}$ approximates the best length-$t$ sequences, which may be generated at the $s$th step (approximated by $\mathcal{B}_{t,s}$) or before the $s$th step (approximated by $\mathcal{A}_{t,s-1}$). Therefore, we can obtain $\mathcal{A}_{t,s}$ by merging $\mathcal{B}_{t,s}$ and $\mathcal{A}_{t,s-1}$:
\begin{align}
    \mathcal{A}_{t,s} = \operatorname{top-}\!K  \Big\{ \mathcal{A}_{t,s-1} \cup \mathcal{B}_{t,s} \Big\} \label{eq:merge_new}
\end{align}
where the $\text{top-}\!K$ operation ranks the sequences by their total scores at and before the $s$th step, given by $ \sum_{s'=t}^{s} u_{s'}(\mathbf b)$. 
This ensures that our approximation algorithm considers the marginalization over $s$ in Eqn.~\eqref{eq:seqmap_obj}. 

Appendix~\ref{app:algo_detail} provides the pseudocode of our decoding algorithm.

\begin{figure}[t]
    \centering
    \includegraphics[width=1.0\linewidth]{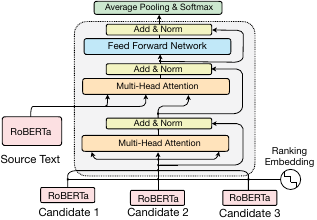}
    \caption{The neural architecture of our reranker. This example assumes a beam size of $K = 3$.}
    \label{fig:reranker}
\end{figure}

\subsection{Reranker}
\label{sec:reranker}
Our SeqMAP beam search finds several high-probability summary candidates,  providing an opportunity to further improve the generation quality using a reranker \cite{och-etal-2004-smorgasbord,lee-etal-2021-discriminative,ravaut-etal-2022-summareranker}. In particular, we train the reranker by predicting which beam candidate has the most overlapped words compared with the groundtruth summary.

\textbf{Neural Architecture.}
Figure~\ref{fig:reranker} depicts the overall architecture of our reranker. 

First, we use the pretrained RoBERTa model \cite{zhuang-etal-2021-robustly} to represent the source text~$\mathbf x$ and each of the summary candidates $\mathbf{b}^{(k)}$, for $1 \leq k \leq K$, individually. 

Then, we build a one-layer Transformer block to predict the best summary for~$\mathbf x$. We start by proposing a novel approach of rank embeddings (initialized randomly and trained by backpropagation) to capture the rank of a candidate given by the DAT model. The rank embedding has the same dimension as RoBERTa, and the $k$th rank's embedding is added to the RoBERTa representation of every word in the $k$th candidate. After that, self-attention is performed among the words in different candidates, and cross-attention is performed to fetch information from the source text. Finally, we pool the representations of all the time steps (including all the words in different candidates) and apply a $K$-way softmax to predict the best candidate.

\textbf{Training Objective.} The target of the classification is the beam candidate $\mathbf{b}^{(k)}$ having the highest number of overlapped words compared with the groundtruth summary $\mathbf{y}$ of a training sample.
We use the standard cross-entropy loss for training, and apply label smoothing \cite{ls} to improve generalization.

\section{Experiments}
\subsection{Setup}
Our experimental setups generally follow the previous non-autoregressive length-control studies \cite{liu-etal-2022-learning,liucharacter}.

\textbf{Datasets.} 
We evaluated the proposed method on the Gigaword headline generation dataset \cite{graff2003english} and the DUC2004 dataset \cite{bommasani-cardie-2020-intrinsic}. The Gigaword dataset pairs news articles with their headlines; it contains 3.0M training pairs, 189K validation pairs, and 1951 test pairs. The DUC2004 dataset contains 500 samples, and is only used as a test set. The performance in the DUC2024 experiment is obtained by the model trained on Gigaword.

\textbf{Metrics.}
We used ROUGE scores \cite{lin-2004-rouge} as the main metric for the quality evaluation of generated summaries. 
Specifically, we reported ROUGE-$n$, for $n \in \{1, 2\}$, to measure the $n$-gram overlaps between the generated summary and the ground truth, and reported ROUGE-L to measure the length of the longest common subsequence.

In addition, we used a large language model (LLM) as a surrogate for human evaluation. We performed pairwise comparisons between the generations of our SeqMAP and baseline models. We reported the win, loss, and tie percentages.

\textbf{Implementation Details.}
For the neural architecture of our DAT models, we used the Transformer-base \cite{attentionisallyouneed} as the backbone. To train the models, we used a batch size of 64K tokens, with a maximum of 100K updates. For regularization, we set the dropout rate to 0.1 and the label smoothing factor to 0.1. For decoding, our SeqMAP had a beam size of $K=20$ and a vocabulary exploration of $V=5$, as mentioned after Eqn.~(\ref{eq:expand_only}).
More details can be found in our GitHub repository (Footnote~\ref{foot:link}), where the configurations of training and inference are included.

\begin{table*}[t]
    \centering
    \resizebox{0.8\textwidth}{!}{%
    \begin{tabular}{lrlrccccd{2.1}} \toprule
    Ratio                 & \#& Model                  & \%Truncate & R-1   & R-2   & R-L   & R-Sum & \multicolumn{1}{c}{Length}  \\ \midrule
    \multirow{7}{*}{20\%} & 1 & AT w/ Truncate           & 60.28\%~\,        & 33.16 & 15.02 & 30.91 & 79.09 & 6.1         \\
                        & 2 & AT-LenInfo w/ Truncate   &  1.03\%~\,         & 33.53 & 14.32 & 31.21 & 79.06 & 5.7        \\
                        & 3 & BERT-CRF w/ Truncate     & 39.98\%~\,        & 31.94 & 13.67 & 30.09 & 75.70 & 5.8         \\
                        & 4 & CTC Length Control       & 0.00\%~\,         & 34.14 & 13.61 & 31.96 & 79.71 & 6.9        \\
                        & 5 & DAT PathMAP            & 0.00\%~\,         & 35.19 & 16.95 & 32.84 & 84.98 & 6.9        \\
                        & 6 & DAT SeqMAP (Ours)       & 0.00\%~\,        & \textbf{36.34} & \textbf{17.29} & \textbf{33.81} & \textbf{87.44} & 6.9        \\ 
                        & 7 & ~~~~ w/o Reranking       & 0.00\%~\,         & 35.63 & 17.24 & 33.28 & 86.15 & 6.8        \\
                        \midrule
    \multirow{7}{*}{25\%} & 1 & AT  w/ Truncate          & 33.52\%~\,        & 34.52 & 15.78 & 31.90 & 82.20 & 6.8        \\
                        & 2 & AT-LenInfo w/ Truncate   &  1.54\%~\,        & 35.12 & 16.19 & 32.60 & 83.91 & 7.3        \\
                        & 3 & BERT-CRF  w/ Truncate   & 18.93\%~\,        & 32.85 & 14.23 & 30.86 & 77.94 & 6.2        \\
                        & 4 & CTC Length Control      & 0.00\%~\,         & 34.51 & 13.89 & 32.08 & 80.48 & 8.5        \\
                        & 5 & DAT PathMAP           & 0.00\%~\,         & 36.11 & 17.43 & 33.41 & 86.95 & 8.5        \\
                        & 6 & DAT SeqMAP (Ours)      & 0.00\%~\,         & \textbf{36.74} & 17.22 & \textbf{33.87} & \textbf{87.83} & 8.4        \\ 
                        & 7 & ~~~~ w/o Reranker      & 0.00\%~\,         & 36.30 & \textbf{17.40} & 33.64 & 87.34 & 8.4        \\
                        \midrule
    \multirow{7}{*}{30\%} & 1 & AT w/ Truncate           & 20.04\%~\,        & 35.07 & 16.00 & 32.35 & 83.42 & 7.1        \\
                        & 2 & AT-LenInfo w/ Truncate   &  4.66\%~\,        & 35.29 & 16.15 & 32.44 & 83.88 & 8.8        \\
                        & 3 & BERT-CRF w/ Truncate     & 9.65\%~\,         & 33.08 & 14.44 & 31.08 & 78.60 & 6.4        \\
                        & 4 & CTC Length Control       & 0.00\%~\,         & 34.04 & 13.63 & 31.49 & 79.16 & 10.0       \\
                        & 5 & DAT PathMAP            & 0.00\%~\,         & 35.57 & 16.65 & 32.88 & 85.10 & 9.9        \\
                        & 6 & DAT SeqMAP (Ours) & 0.00\%~\,         & \textbf{36.24} & 16.74 & \textbf{33.34} & \textbf{86.32} & 9.9        \\ 
                        & 7 & ~~~~ w/o Reranker       & 0.00\%~\,         & 35.80 & \textbf{16.76} & 33.04 & 85.60 & 9.9        \\
                        \bottomrule
    \end{tabular}}
    \caption{Results on the Gigaword dataset, where we set the length constraint to be 20\%, 25\%, and 30\% of the source length. \%Truncate is the percentage of sentences that require truncation to meet the length requirement. R-1, R-2, R-L, and R-Sum denote the ROUGE-1, ROUGE-2, ROUGE-L, and the sum of the three ROUGE scores, respectively. }
    \label{tab:main}
\end{table*}

\subsection{Results and Analyses}
\textbf{Main Results on Gigaword.} 
Table~\ref{tab:main} presents our main results on the Gigaword dataset.
For thorough comparison, we evaluate the approaches in three scenarios: setting the summary lengths to 20\%, 25\%, and 30\% of the source text length. For non-strict length-control methods, the output will be truncated if the length exceeds the limit. 

We first include two baseline methods that do not have length control: 1) the standard autoregressive Transformer \cite[AT,][]{attentionisallyouneed}, and 2) BERT-CRF \cite{su-etal-2021-non}, which is a non-autoregressive method that uses BERT as the encoder \cite{devlin-etal-2019-bert} and applies a conditional random field \cite[CRF,][]{crf} for decoding. 
As seen in Table~\ref{tab:main}, a large portion of the generated sentences from AT and BERT-CRF require truncation to meet the length requirement (Rows 1 and 3), resulting in incomplete summaries. Also, their ROUGE scores are low in general.

We further consider a soft length-control method: AT-LenInfo \cite{liu-etal-2018-controlling}, which integrates a length embedding to softly guide the output length of the autoregressive Transformer. As we can see, AT-LenInfo largely reduces the truncation ratio, while achieving similar ROUGE scores to AT.  However, AT-LenInfo still has 1.03--4.66\% of sentences truncated, suggesting that such a soft-constraint method is inadequate for length control.

On the other hand, the CTC \cite{liu-etal-2022-learning} and our DAT methods perform exact length control by dynamic programming; thus, no truncation is required (Rows 4--7). Among these models, our DAT (with any inference method) consistently outperforms CTC, verifying that DAT is a more powerful non-autoregressive model.

We also observe that our SeqMAP method, with or without the reranker, is better than PathMAP, with only one exception (the R-2 score in the 25\% setting); our reranker further improves the total ROUGE scores (R-Sum) by 0.49--1.32 points. The results verify the superiority of our SeqMAP objective and the effectiveness of our decoding algorithms. 

\begin{table}[t]
\centering
    \resizebox{0.48\textwidth}{!}{%
\begin{tabular}{lcccc} \\ \toprule
                          & R-1   & R-2   & R-L    & R-Sum      \\ \midrule
AT w/ Truncate            & 27.03 & 8.92  & 24.20  &  60.15     \\
AT-LenInfo w/ Truncate    & 27.88 & 9.28  & 25.58  &  62.74     \\
BERT-CRF w/ Truncate      & 23.69 & 7.16  & 21.64  &  52.49     \\
CTC  Length Control       & 28.67 & 8.29  & 26.34  &  63.30     \\
DAT PathMAP               & 28.77 & 9.98  & 26.48  &  65.23     \\
DAT SeqMAP (Ours)         & \textbf{30.11} & \textbf{10.29} & \textbf{27.22} &  \textbf{67.62}  \\
~~~ w/o Reranker         & 29.17 & 10.27 & 26.66  &  66.10     \\
                        \bottomrule
\end{tabular}}
\caption{Results on the DUC2004 dataset. The summary length is set to 20\% of the source length.}
\label{tab:duc2004}
\end{table}

\begin{table}[t]
\centering
\resizebox{0.42\textwidth}{!}{%
\begin{tabular}{@{}l@{}}
\toprule
Given the text:  \textbf{[Source]} \\ 
Consider two summaries of the text: \\
Summary \textbf{[ID1]}: \textbf{[Summary1]} \\
Summary \textbf{[ID2]}:  \textbf{[Summary2]} \\
\parbox[t]{0.48\textwidth}{A good summary is a shorter piece of text that has the essence of the original and adheres to coherence, faithfulness, relevance, and overall quality as defined above. Which summary is better?} \\
Answer:  \\
\bottomrule
\end{tabular}}
\caption{Our prompt template for LLM-based pairwise evaluation. Here, ``\textbf{Source}'' is the text to be summarized. The choices of IDs are ``1'' and ``2''; ``\textbf{Summary1}'' and ``\textbf{Summary2}'' are substituted with model-generated text. Since LLM is not robust to candidate ID and order~\cite{zheng2023large,shen-etal-2023-large}, we enumerate different combinations for a given case, resulting in four LLM queries.}
\label{tab:llm_prompt}
\end{table}

\begin{table}[t]
\centering
\resizebox{0.48\textwidth}{!}{%
\begin{tabular}{lcrr}
\toprule
Pairwise Comparison                 & Win  & Loss~~\, & Tie~~~~  \\ \midrule
Ours vs. AT w/ Trunc.            & \textbf{68.70\%}       & 0\%           & 31.30\%       \\
Ours vs. AT-LenInfo w/ Trunc. \!\!  & \textbf{69.45\%}       & 0\%           & 30.55\%       \\
Ours vs. CTC Length Control        & \textbf{72.40\%}       & 20.00\%       & 7.60\%        \\
Ours vs. DAT PathMAP              & \textbf{43.77\%}       & 32.04\%       & 24.18\%       \\ \bottomrule
\end{tabular}
}
\caption{LLM pairwise comparison between DAT SeqMAP and baseline methods. For each sample, 
we determine the outcome of ``win,'' ``loss,'' or ``tie'' based on the votes of the four LLM queries (by varying candidate order and ID tokens). We report the ratios of wins, losses, and ties of our model.}
\label{tab:llm_eval}
\end{table}

\textbf{Main Results on DUC2004.}
We report the performance on the DUC2004 dataset in Table~\ref{tab:duc2004}.
As seen, the trends mirror those observed on the Gigaword dataset. In particular, our SeqMAP consistently outperforms other baselines in all metrics. 
Since the DUC2004 results are obtained from models trained on Gigaword, we conclude that our approach is transferable to different testing scenarios.

\textbf{LLM Evaluation.}
In addition to ROUGE scores, we prompt a large language model (LLM), in particular, \texttt{gpt-4-0125-preview}, serving as a surrogate for human evaluation. Concretely, we perform pairwise comparison between the outputs of our SeqMAP and baseline models on the Gigaword test set. For each comparison, we query an LLM four times by enumerating the order and IDs of the candidates for better robustness; the exact prompt is presented in Table~\ref{tab:llm_prompt}.

Table~\ref{tab:llm_eval} shows the results of LLM evaluation. We first observe that our SeqMAP dominates AT, AT-LenInfo, and CTC baselines with a winning rate of 68.80--72.40\%. 
Further, our SeqMAP has an 11.73\% higher winning rate than PathMAP with the same DAT backbone.
Overall, our pairwise LLM evaluation is consistent with the comparison based on ROUGE scores (Tables~\ref{tab:main} and \ref{tab:duc2004}), further verifying the effectiveness of our SeqMAP approach.

\begin{table}[t]
    \centering
    \resizebox{0.48\textwidth}{!}{%
    \begin{tabular}{llcc} \toprule
    \#  &   Model       & Word Novelty & Reordering Degree \\ \midrule
    1   &   CTC         & 18.41\% & 8.69\%       \\
    2   &   DAT PathMAP & 21.80\% & \textbf{9.09\%}        \\ 
    3   &   DAT SeqMAP (Ours)      & \textbf{22.48\%} & 9.01\%       \\ \midrule
    4   &   Human Reference   & 50.51\% & 8.05\%        \\
    \bottomrule
    \end{tabular}}
    \caption{The word novelty and reordering ratio of CTC, DAT PathMAP, DAT SeqMAP, and the human reference. The results are based on the Gigaword test set.}
    \label{tab:CTC_vs_DAT}
\end{table}

\textbf{Generation Novelty.} In this work, we hypothesize that the Directed Acyclic Transformer \cite[DAT,][]{huang2022directed} is more flexible in predicting words and determining their order, compared with the Connectionist Temporal Classification (CTC) model \cite{graves2006connectionist}, which is used in \newcite{liu-etal-2022-learning} for length-control summarization. To verify this hypothesis, we {adopt} two metrics, namely \emph{word novelty} and the \emph{reordering degree}, to compare the generated summaries by DAT and~CTC.

In particular, word novelty measures the percentage of the summary's words that are not in the source text, indicating the method's flexibility in word choice. 
The reordering degree assesses the flexibility in generation order. It is computed by the following steps: 1) Align the words in the summary to the source text,\footnote{We use FastAlign \cite{dyer-etal-2013-simple}, a tool for word alignment, to obtain the target-to-source alignments.} 2) Enumerate every pair of word positions $(i,j)$ such that $i<j$ in the source, and 3) Count the fraction of the order being reversed in the output.

Table~\ref{tab:CTC_vs_DAT} presents the results on the Gigaword dataset. Interestingly, humans tend to use novel words but preserve the word order. The DAT model has the highest reordering degree, which, although not agreeing with humans, verifies our hypothesis that DAT is more flexible than CTC in word ordering. In addition, DAT also yields more novel words than CTC.

\begin{table}[t]
    \centering
    \resizebox{.85\linewidth}{!}{$
    \begin{tabular}{llcr} \toprule
    \#  &   Model              & Sentences/s    & Words/s\!         \\ \midrule
    1   &   AT  w/ Trunc.         & 10.59          & 90.99           \\
    2   &   CTC Length Control                & 25.61          & 219.71          \\
    3   &   DAT  PathMAP        & \textbf{61.30} & \textbf{520.64} \\
    4   &   DAT SeqMAP w/ Reranker           & 10.97          & 104.08          \\ 
    5   &   DAT SeqMAP w/o Reranker & 20.75          & 177.00          \\ \bottomrule
    \end{tabular}$}
    \caption{Inference speed of different methods.}
    \label{tab:speed}
\end{table}

\textbf{Inference Speed.}
Non-autoregressive (NAR) models are originally designed to improve inference efficiency for text generation \cite{gu2018nonautoregressive,gu-kong-2021-fully,huang2022non}. In this work, we compare the inference speed following the convention of NAR research, where the batch size is set to 1, mimicking real-world scenarios where user queries come one after another.

As seen from Table~\ref{tab:speed}, NAR-based methods, except DAT SeqMAP with the reranker (Row~4), are faster than the autoregressive baseline (Rows 2, 3, and 5 versus Row 1). It is understandable that our DAT SeqMAP with the reranker is not as efficient as other NAR models because it requires a dynamic programming-based beam search and an external neural module for reranking. Nevertheless, our method is still faster than the autoregressive model, showing its practical value in application. 

It is worth mentioning that our SeqMAP implementation has not been optimized for parallelism yet, as opposed to the PathMAP implementation~\cite{shao-etal-2022-viterbi}. Therefore, there is room to further improve the efficiency of our SeqMAP.

\textbf{Decoding Hyperparameters.}
Our SeqMAP decoding method has two hyperparameters controlling its search scope: 
\begin{compactitem}
    \item $K$: the beam size for the candidate set $\mathcal{A}_{s,t}$ as in Eqn.~\eqref{eq:beam_a}, and
    \item $V$: the number of words explored at each step, as mentioned after Eqn.~\eqref{eq:expand_only}.
\end{compactitem}
We analyzed their impact by varying $K$ (with fixed $V=5$) and varying $V$ (with fixed $K=20$). The results are listed in Tables~\ref{tab:k_m} and \ref{tab:k_v}, respectively.

First, we observe that SeqMAP's performance generally improves as $K$ or $V$ increases, with or without the reranker. This shows that our approximation algorithm for SeqMAP benefits from increasing its search scope.

When $K$ or $V$ is increased beyond a certain value ($K=20$ or $V=5$), the performance stops increasing, or even slightly decreases in some metrics such as R-2. This is consistent with autoregressive beam search, where a moderately sized beam works the best~\cite{stahlberg-byrne-2019-nmt,meister-etal-2020-beam,wen2024ebbs}.  

\begin{table}[t]
    \centering
    \resizebox{0.48\textwidth}{!}{
    \begin{tabular}{llccccc} \toprule
                            & Model                          & $K$ & R-1   & R-2   & R-L   & R-Sum \\ \midrule
    \multicolumn{2}{c}{PathMAP}                             & -    & 36.11 & 17.43 & 33.41 & 86.95 \\ \midrule
    \multirow{8}{*}{\rotatebox[origin=c]{90}{SeqMAP}} & \multirow{4}{*}{w/o Reranker}        & 10   & 36.27 & 17.3  & 33.58 & 87.15 \\
                            &                                & 15   & 36.27 & 17.34 & 33.59 & 87.20  \\
                            &                                & 20   & 36.30 & 17.40 & 33.64 & 87.34 \\
                            &                                & 25   & 36.27 & 17.36 & 33.6  & 87.23 \\ \cmidrule{2-7}
                            & \multirow{4}{*}{w/ Reranker} & 10   & 36.61 & 17.27 & 33.8  & 87.68 \\ 
                            &                                & 15   & 36.65 & \textbf{17.41} & 33.80 & 87.86 \\
                            &                                & 20   & \textbf{36.74} & 17.22 & \textbf{33.87} & 87.83 \\
                            &                                & 25   & \textbf{36.74} & 17.32 & \textbf{33.87} & \textbf{87.93} \\ \bottomrule
    \end{tabular}}
    \caption{Reranker performance of different $K$. Here, $V=5$, and the length ratio is 0.25.}
    \label{tab:k_m}
\end{table}

\begin{table}[t]
    \centering
    \resizebox{0.48\textwidth}{!}{%
    \begin{tabular}{llccccc} \toprule
                            &                             & $V$ & R-1   & R-2   & R-L   & R-Sum \\ \midrule
    \multicolumn{2}{c}{PathMAP}                           & -     & 36.11 & 17.43 & 33.41 & 86.95 \\ \midrule
    \multirow{10}{*}{\rotatebox[origin=c]{90}{SeqMAP}} & \multirow{5}{*}{w/o Reranker}  & 1     & 36.23 & 17.37 & 33.59 & 87.19 \\
                            &                             & 3     & 36.24 & 17.34 & 33.58 & 87.16 \\
                            &                             & 5     & 36.30 & 17.40 & 33.64 & 87.34 \\
                            &                             & 7     & 36.24 & 17.34 & 33.59 & 87.17 \\
                            &                             & 9     & 36.24 & 17.34 & 33.59 & 87.17 \\ \cmidrule{2-7}
                            & \multirow{5}{*}{w/ Reranker} & 1     & 36.45 & 16.92 & 33.62 & 86.99 \\
                            &                             & 3     & 36.73 & 17.16 & 33.84 & 87.73 \\
                            &                             & 5     & \textbf{36.74} & \textbf{17.22} & \textbf{33.87} & \textbf{87.83} \\
                            &                             & 7     & \textbf{36.74} & 17.20  & 33.85 & 87.79 \\
                            &                             & 9     & \textbf{36.74} & 17.18 & 33.85 & 87.77 \\ \bottomrule
    \end{tabular}}
    \caption{Reranker performance of different $K$. Here, $V=20$, and the length ratio is 0.25.}
    \label{tab:k_v}
\end{table}

\begin{table}[t]
    \centering
    \resizebox{0.48\textwidth}{!}{%
    \begin{tabular}{lccccccc} \toprule
    \#  & LS & RankEmb & Pretrain & R-1    & R-2         &  R-L      & R-Sum \\ \midrule
    1   & \Checkmark  & \Checkmark      & \Checkmark        & \textbf{36.61} & 17.27      &   \textbf{33.80}  & \textbf{87.68}              \\
    2   &             & \Checkmark      & \Checkmark        & 36.50          & 17.18      &   33.73  & 87.41               \\
    3   & \Checkmark  &                 & \Checkmark        & 36.23          & 17.28      &   33.57  & 87.08               \\ 
    4   & \Checkmark  & \Checkmark      &                   & 36.38          & \textbf{17.34}      &   33.68  & 87.41               \\ \bottomrule
    \end{tabular}}
    \caption{Ablation study on the effect of label smoothing, positional embedding, and pretraining. To save the training time, we set $K$ to 10 and $V$ to 5.}
    \label{tab:reranker}
\end{table}

\textbf{Analysis of the Reranker.}
As mentioned in Section~\ref{sec:reranker}, our neural reranker for summarization consists of several key components: 1) \emph{Label smoothing}, which is used for regularization \cite{ls}, 2) \emph{Rank Embedding}, which captures the rank of a candidate given by DAT, and 3) \emph{Pretraining}, where we use RoBERTa-base \cite{zhuang-etal-2021-robustly} for sentence representation. We perform an ablation study on the impact of the three components, and present the results in Table~\ref{tab:reranker}. 

As seen, removing label smoothing results in a 0.27 decrease in R-Sum (Rows 1 and 2). This is because the difference between beam candidates can be minimal, and lowering the confidence of the reranker in training leads to better generalization.

When removing the rank embedding, we observe a 0.6 decrease in R-SUM (Rows 1 and 3). 
This confirms our intuition that incorporating the DAT model's ranking as prior knowledge is beneficial.
Additionally, we observe that having Rank Embedding leads to faster training convergence.

Without using the pretrained RoBERTa model (i.e., randomly initializing the weights), R-SUM decreases by 0.27. This drop is anticipated since the pretrained model enhances text understanding.

\section{Related Work}
Non-autoregressive (NAR) models predict words independently, 
and are initially developed to increase the inference speed of neural machine translation \cite{gu2018nonautoregressive,lee-etal-2018-deterministic,qian2020glancing,gu-kong-2021-fully,huang-etal-2023-multilingual}.
Recently, NAR models have been adapted for length-control summarization in our previous work \cite{liu-etal-2022-learning,liucharacter}, where we find that NAR's independent word predictions allow the length-control tasks to be divided into several independent sub-tasks, resulting in an efficient exploration of the NAR output space. Therefore, we have developed dynamic programming algorithms based on the Connectionist Temporal Classification (CTC) model \cite{graves2006connectionist}.

Our work introduces a novel decoding algorithm that leverages the Directed Acyclic Transformer \cite[DAT,][]{huang2022directed}. Unlike CTC, which preserves the order of the source sequence \cite{chuang-etal-2021-investigating,shao2022}, DAT offers greater flexibility in word selection and generation order. While recognizing the value of existing DAT-based decoding methods \cite{shao-etal-2022-viterbi} for managing length, we identify their limitations and propose a new SeqMAP approach.

Our proposed reranker is inspired by the reranking methods in machine translation \cite{och-etal-2004-smorgasbord,lee-etal-2021-discriminative} and summarization \cite{ravaut-etal-2022-summareranker}, where a list of $n$-best sequences are presented to an external model for scoring. 

\section{Conclusion}
This work proposes a novel SeqMAP decoding method based on the Directed Acyclic Transformer (DAT) for length-control summarization. Our \mbox{SeqMAP} is a beam search-based dynamic programming algorithm, which bridges the gap between the training and inference of DAT. Experimental results on the Gigaword and DUC2004 datasets demonstrate the effectiveness of our SeqMAP approach.

\section{Limitations}
One potential limitation of this work is that the proposed {algorithm for} SeqMAP does not guarantee to find the best sequence in DAT's decoding space. As discussed, an exact algorithm may not exist due to the computational complexity of the Maximum A Posteriori (MAP) problems \cite{koller2009probabilistic,de2011new}.
Nevertheless, we propose an approximate algorithm for SeqMAP and empirically show that it is consistently better than PathMAP. 

We also notice a slight performance decrease with a large scope of our beam search (controlled by $K$ and $V$ mentioned in Section~\ref{sec:beam}). This phenomenon, previously observed in autoregressive text generation models \cite{stahlberg-byrne-2019-nmt,meister-etal-2020-beam}, might stem from the label bias issue \cite{crf,huang-etal-2021-globally}. We leave the investigation of this phenomenon as future work.

\section*{Acknowledgments}
We would like to thank all reviewers and chairs for their comments.
This research was supported in part by the Natural Science Foundation of China under Grant No.\,62376133.
This research was also supported in part by the Natural Sciences and Engineering Research Council of Canada (NSERC) under Grant Nos. RGPIN-2020-04440 and RGPIN-2020-04465, 
the Amii Fellow Program, the Canada CIFAR AI Chair Program, the Alberta Innovates Program, the Digital Research Alliance of Canada (alliancecan.ca), and a donation from DeepMind. 

\bibliography{custom} 

\appendix
\section{The Pseudocode for Our Algorithm}
\label{app:algo_detail}

We present in Algorithm~\ref{algo:seqmap} the pseudocode of our beam search-based dynamic programming algorithm for the SeqMAP objective. 

As a recap of Section~\ref{sec:beam}, $\mathcal A_{t,s}$ is the approximated \topK{K} length-$t$ sequences that are generated \emph{at or before} DAT's $s$th step, whereas $\mathcal B_{t,s}$ is the approximated \topK{K} length-$t$ sequences generated \emph{exactly at} the $s$th step. 


In Algorithm~\ref{algo:seqmap}, Lines 2--4 are the initialization of $\mathcal A_{0,s}$, which is set to be the special token $\langle\text{bos}\rangle$  indicating the beginning of a sentence, and the score for the token is $1$.

Lines 5--14 are the iterations for computing the dynamic programming (DP) table for each $t \in \{ 1, \cdots, T \}$ and $s \in \{1, \cdots, S \}$. Within each iteration,
\begin{itemize}
    \item Lines 7--12 correspond to our $\textsc{Expand}$ operation, which approximates $\mathcal B_{t,s}$ by expanding sequences in $\mathcal A_{t-1,s-1}$. Specifically, Line 10 corresponds to the sequence expansion as described in Eqn.~\eqref{eq:expand_only}, and Line 11 computes the scores of newly generated sequences as described in Eqn.~\eqref{eq:expand_score}. Then, the \topK{K} operation in Line 12 prevents the beam from growing exponentially.
    \item Lines 13--14 correspond to the $\textsc{Merge}$ operation, which 
    combines the newly formed sequences in $\mathcal B_{t,s}$ and the inherited sequences from $\mathcal A_{t,s-1}$. Again, we have the \topK{K} operation to keep the beam search tractable, where the ranking is given by summing the probabilities over different steps, as explained after Eqn.~\eqref{eq:merge_new}.
\end{itemize}

Our DP algorithm terminates when we have computed $\mathcal A_{T,S}$, which contains a few length-$T$ sequences of high probability given by DAT. Finally, Line 15 applies a reranker (as described in Section~\ref{sec:reranker}) to determine the best summary in $\mathcal A_{T,S}$.

\let\oldnl\nl
\newcommand{\nonl}{\renewcommand{\nl}{\let\nl\oldnl}}
\begin{algorithm*}
\RestyleAlgo{ruled}
\SetKwComment{Comment}{$\triangleright$ }{ }
\SetAlgoLined
\DontPrintSemicolon
\caption{Our beam search-based dynamic programming algorithm for the SeqMAP objective} \label{algo:seqmap}
\textbf{Input:} $S$: total prediction steps in DAT;\quad  $\mathcal V$: vocabulary;\quad $l_{i,j}$: link probability for $i,j \in \{ 1, \cdots, S \}$;\quad $w_{s,\mathrm v}$: word probability for $s \in \{1, \cdots, S\}$ and $\mathrm v \in \mathbf{V}$;\quad $K$: beam size;\quad\quad $V$: number of words to be expanded;\quad $T$: the desired length \\ 
\Comment{{Initialization}}: 
\For{ $s := 1, \cdots, S$}{   
    $\mathcal A_{0, s} :=  \{ \langle \text{bos} \rangle \} $ ~~~~
    \Comment{$\langle\text{bos}\rangle$ is the starting token}
    ${u_{s} (\langle \text{bos} \rangle)}  := 1 $  ~~~~~~   \Comment{$\bm u_s (\mathbf b)$ is the score of generating $\mathbf b$ at step $s$}
} 
\Comment{{Recursive steps}} 
\For{ $t := 1, \cdots, T $}{
    \For{$s := t, \cdots, S$}{
        \Comment{\textit{{Expand}} $\mathcal A_{t-1,s-1}$ to obtain $\mathcal B_{t,s}$}
        \For{{\normalfont each $\mathbf b$ in $\mathcal A_{t-1,s-1}$}}{
            $B_{t, s} := \{ \}$  ~~~~~~~~~ \Comment{Initialization}
            \For{{\normalfont each $\operatorname{top-}\!V$ most probable predicted words $\mathrm v'\in\mathcal V$ at step $s$}
            }{
                $\mathcal B_{t, s} :=  \mathcal B_{t, s} \cup  \{\mathbf b \oplus {\mathrm  v^\prime}\} $
                ~~~~~~~~~~~~ \Comment{$\oplus$ denotes string concatenation} 
                ${u_{s}(\mathbf b \oplus {\mathrm v^\prime})}  :=  w_{s,\mathrm v^\prime} \sum_{s'=t-1}^{s-1} u_{s'}(\mathbf b) \cdot  l_{s',s}$ ~~~~ \Comment{Marginalization}
            }
            $\mathcal B_{t, s} := {\operatorname{top-}\!K}~ \big\{ \mathcal B_{t,s} \}$ 
            \quad\quad\quad\quad\quad~ \Comment{Ranking is based on $u_s(\mathbf b)$}  
        }
        \Comment{\textsc{Merge} $\mathcal B_{t,s}$ and  $\mathcal A_{t,s-1}$} 
            \If {$s > t$}{
                $\mathcal A_{t,s} := \operatorname{top-}\!K\big\{ \mathcal B_{t,s} \cup \mathcal A_{t,s-1} \big\}$ 
        \quad~~~\Comment{Ranking is based on $\sum_{s'=t}^{s} u_{s'}(\mathbf b)$} 
        }
    }
}
\Comment{Reranking}
\KwRet the top reranked $\mathbf b$ in $\mathcal A_{T,S}$ according to our reranker (Section~\ref{sec:reranker})
\;
\end{algorithm*}

\end{document}